\documentclass[letterpaper]{article} 
\usepackage{amsmath}
\usepackage{amssymb}
\usepackage{times}
\usepackage{bm}
\usepackage{xargs}
\usepackage{multirow}
\usepackage{amsfonts}
\usepackage{color}
\usepackage{booktabs}

\newcommandx\includeImageLineWidth[2][1=1.0]{\includegraphics[width=#1\linewidth]{#2}}

\newcommand{\bt}{{\bm{t}}}

\newcommand{\bv}{{\bm{v}}}

\newcommand{\bM}{{\bm{M}}}

\usepackage{array}
\newcommand{\PreserveBackslash}[1]{\let\temp=\\#1\let\\=\temp}
\newcolumntype{C}[1]{>{\PreserveBackslash\centering}p{#1}}
\newcolumntype{R}[1]{>{\PreserveBackslash\raggedleft}p{#1}}
\newcolumntype{L}[1]{>{\PreserveBackslash\raggedright}p{#1}}

\include{math_commands}
\usepackage{aaai23}  
\usepackage{times}  
\usepackage{helvet}  
\usepackage{courier}  
\usepackage[hyphens]{url}  
\usepackage{graphicx} 
\usepackage{natbib}  
\usepackage{caption} 
\usepackage{algorithm}
\usepackage{algorithmic}
\usepackage{subfiles}
\usepackage{xcolor}
\usepackage{subfigure}
\usepackage{enumitem}
\usepackage{pifont}

\usepackage{newfloat}
\usepackage{listings}
\DeclareCaptionStyle{ruled}{labelfont=normalfont,labelsep=colon,strut=off} 
\floatstyle{ruled}
\newfloat{listing}{tb}{lst}{}
\floatname{listing}{Listing}
\nocopyright

\title{Text-to-Image Generation via Implicit Visual Guidance and Hypernetwork}
\author{
    Xin Yuan,\textsuperscript{\rm 1}
    Zhe Lin,\textsuperscript{\rm 2}
    Jason Kuen,\textsuperscript{\rm 2}
    Jianming Zhang,\textsuperscript{\rm 2}
    John Collomosse\textsuperscript{\rm 2, \rm 3}
}
\affiliations{
    \textsuperscript{\rm 1}University of Chicago, 
    \textsuperscript{\rm 2}Adobe Research
    \textsuperscript{\rm 3}University of Surrey

    yuanx@uchicago.edu, \{zlin, kuen, jianmzha\}@adobe.com, j.collomosse@surrey.ac.uk
}

\newcommand{\cmark}{\ding{51}}%
\newcommand{\xmark}{\ding{55}}%

\begin{document}

\maketitle
\begin{abstract}
We develop an approach for text-to-image generation that embraces additional retrieval images, driven by a combination of implicit visual guidance loss and generative objectives. 
Unlike most existing text-to-image generation methods which merely take the text as input, our method dynamically feeds cross-modal search results into a unified training stage, hence improving the quality, controllability and diversity of generation results. 
We propose a novel hypernetwork modulated visual-text encoding scheme to predict the weight update of the encoding layer, enabling effective transfer from visual information (e.g. layout, content) into the corresponding latent domain.
Experimental results show that our model guided with additional retrieval visual data outperforms existing GAN-based models.
On COCO dataset, we achieve better FID of $9.13$ with up to $3.5 \times$ fewer generator parameters, compared with the state-of-the-art method. 
\end{abstract}

\section{Introduction} \label{sec:intro}
In recent years, Generative Adversarial Networks (GANs)~\cite{goodfellow2014generative} have produced realistic results on the text-to-image generation task~\cite{Xu18,Han17,Han17stackgan2,DBLP:journals/corr/abs-2008-05865,hinz2020semantic, DBLP:conf/cvpr/0010KBLY21}. Despite the success, unstructured text descriptions usually make it difficult for the generator to learn a good distribution coverage.
\begin{figure}[tb]
 \begin{center}
   \subfigure[Difference between our method and most existing works.]{
      \includegraphics[width = \columnwidth]{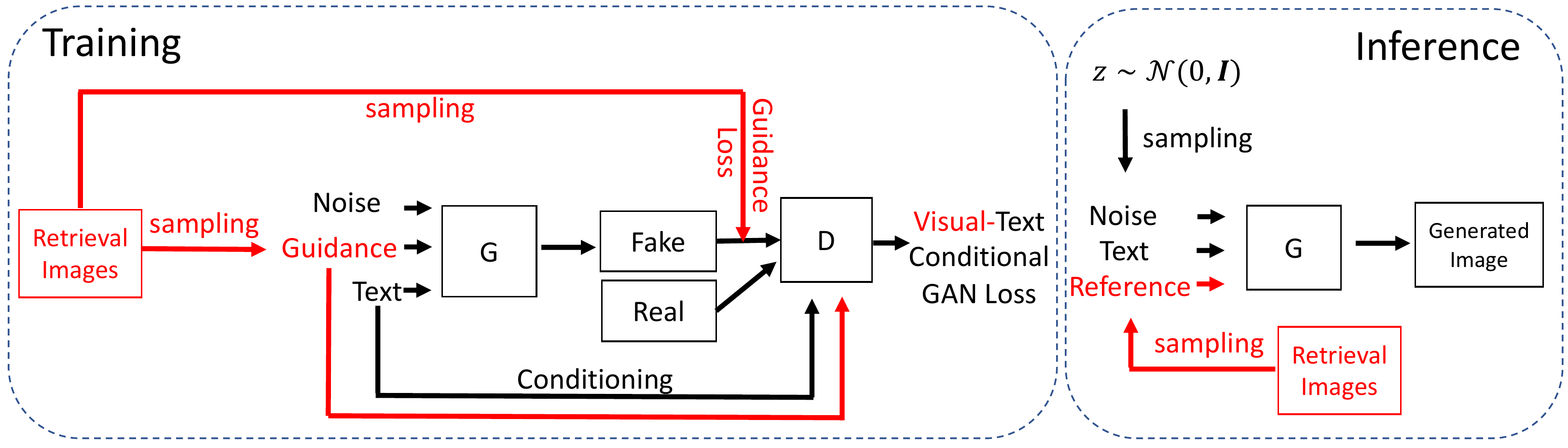}
   }
\end{center}
  \hfill
  \begin{center}
   \subfigure[Scenes generated by our method. \textit{\textbf{left}:} for the \textit{upper} reference with close view of multiple players in the field, generations also present a close view of multiple players; for the \textit{upper} reference with distant view of single player and the whole field, our generations present similar layouts, with audience included. 
\textit{\textbf{right}:} for the \textit{upper} reference with large portion of green vegetables, our generations consistently produce `pizza' images with similar ingredients; for the \textit{bottom} reference with less greens, the generations present a similar content.] {
      \includegraphics[width = \columnwidth]{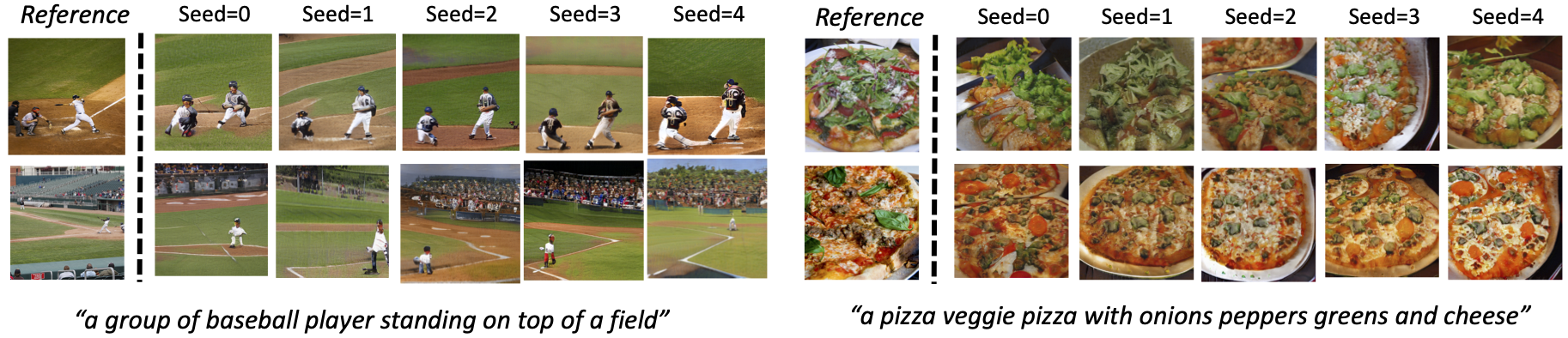}
   }
   \end{center}
   \caption{
     Main idea of the proposed method. (a) Difference between our method and others. As highlighted in {\color{red}{red}}, our method leverages the additional information from retrieval image to dynamically create image-text pairs, which produces more diversified joint distribution and enables better training with additional, easy-acquired visual conditions and guidance. Without any inference-time latent optimization, the model can directly generate high-quality images with diverse and complex scenes using the text description  (b) Even with noise vector fixed (each column), the diversity can still be maintained with much better controllability (each row), as demonstrated by different retrieval references with rich visual information (e.g. styles, layouts, colors and contents).
     (\textit{best view in color.})
   }
   \label{fig:motivation}
\end{figure}
The cross-modal gap between texts and inadequate real image samples during training presents optimization difficulties (e.g. overfitting~\cite{hinz2020semantic, objgan19}, mode collapse), yielding image results with low quality and diversity, especially in the case of generating a complex scene conditioned on one-sentence description. 
XMC-GAN~\cite{DBLP:conf/cvpr/0010KBLY21} tries to bridge the gap by forcing strong text-image correspondence through a contrastive discriminator and achieves state-of-the-art results. However, as shown in Figure~\ref{fig:motivation}(a), it is still trained with fixed pairs labeled by datasets. To fully exploit the intrinsic data properties, long epochs and large model parameters are necessary for cross-modal contrastive loss, requiring huge training resources.
This dilemma can be relieved by training the generator on large-scale visual-language datasets with massive image-text pairs provided ~\cite{DBLP:conf/acl/SoricutDSG18}.
Yet, the fine-grained image captioning efforts on such a large dataset are usually prohibitively heavy.

Our objective is to leverage the power of cross-modal search by feeding text-image pairs that are dynamically created, providing more informative and implicit guidance during training.
The framework design should have following essential properties: 
1) Large capacity to learn cross-modal distributions from dynamic image-text pairs; 
2) Effective encoding scheme for implicit information transfer;
3) Good quality, diversity and controllability at inference time;
4) Can easily leverage samples from  external datasets as guidance.
We aim to learn high-quality text-to-image generation results with additional guidance of easily acquired text-to-image search results. 

Several recently proposed methods~\cite{HinzHW19, hinz2020semantic, casanova2021instanceconditioned, li2022memorydriven,trecs2020} also focus on generating high-quality images using additional information.
For example, 
IC-GAN~\cite{casanova2021instanceconditioned} extends and improves unconditional GANs by modeling the neighborhood distribution of an instance feature. 
Yet, jointly encoding multimodal information in a unified training framework is not the focus of IC-GAN. Given a text description, IC-GAN requires an inference-time CLIP-guided~\cite{DBLP:conf/icml/RadfordKHRGASAM21} noise optimization for image generation.
OP-GAN~\cite{hinz2020semantic} simultaneously learns the layout information provided by individual objects and generates an image conditioned on both layouts and text.
However, OP-GAN requires multiple training stages and more fine-gained object bounding boxes so it is not efficient.
MemoryGAN~\cite{li2022memorydriven} builds a retrieval image memory bank and then selectively feeds image features for guidance at each stage of the network. 
However, even with both global and region features extracted, it still requires multi-stage discriminators~\cite{Xu18} to generate high-resolution images (e.g. $256 \times 256$), suggesting the memory features are not fully exploited by the encoding stage.


In this paper, we take a unified view of implicit visual guidance and generative objectives and develop a new text-to-image generation framework, as shown in Figure~\ref{fig:motivation}.
To be specific, we first conduct offline cross-modal search to build up candidate image retrieval results. 
A hypernetwork architecture is used in the text-conditioned encoding layer to predict the weights updating and effectively transfer visual information (e.g. content, layout) from retrieval results into the latent representation, hence improving the controllability of both generator and discriminator training via back-propagation.
During inference, given a text description, the generation diversity can be ensured via not only random noise vectors, but also candidate visual feature sampling in a more controllable manner.
We summarize our contributions as two-fold:
\begin{itemize}[leftmargin=.15in]
   \item{%
      \textbf{Unified Training.}~
      Our training framework simultaneously encodes rich information from text description and cross-modal search results in a unified procedure with carefully designed visual-text conditional mapping layer and implicit visual guidance loss. In addition, as shown in Figure~\ref{fig:motivation} and~\ref{fig:framework}, our framework randomly samples candidate images with respect to text description. It has the flexibility to incorporate any new external image source. 
   }%
   \item{%
      \textbf{Better quality, controllability and diversity.}~
      The generator trained by our method can generate high-quality images adhere to semantic information in text descriptions.
      Instead of merely relying on noise sampling, a more controllable and diversified inference is reached by accommodating external features that share similar cross-modal similarity as reference.
      The model achieves excellent quantitative and qualitative results under common evaluation protocols. 
   }%
\end{itemize}
We demonstrate these advantages through comparing experimental results with recent GAN-based methods on CUB~\cite{WahCUB_200_2011} and the challenging COCO~\cite{LinMBHPRDZ14} datasets.

\section{Related Work}
\textbf{Direct text-guided image generation.}
~\cite{Xu18,Han17,Han17stackgan2,DBLP:journals/corr/abs-2008-05865,hinz2020semantic, DBLP:conf/cvpr/0010KBLY21, zhu2019dm} are proposed by only using the text descriptions as conditions for GANs. 
For example, StackGAN~\cite{Han17,Han17stackgan2} progressively generates images in a coarse-to-fine manner.
~\cite{Xu18} improves ~\cite{Han17stackgan2} with cross-modal attention mechanism.
DM-GAN~\cite{zhu2019dm} proposes a dynamic memory module with both reading and writing gates to refine image content.
XMC-GAN~\cite{DBLP:conf/cvpr/0010KBLY21} explores the multimodal contastive losses in a simple one-stage GAN framework and achieves the state-of-the-art performance in terms of image quality and image-text alignment. However, it still learns the joint distribution over fixed and insufficient text-image pairs provided by datasets.


\noindent \textbf{GANs with additional information.}
~\cite{HinzHW19, hinz2020semantic, casanova2021instanceconditioned, li2022memorydriven,trecs2020} use additional information for generating high-quality images.
For example, 
Recently proposed IC-GAN~\cite{casanova2021instanceconditioned} provides extensions to unconditional GANs by feeding additional instance feature's neighbors for better distribution learning.
OP-GAN~\cite{HinzHW19} generates images conditioned on both layouts and text simultaneously, followed by a separate layout learning scheme from individual objects.
MemoryGAN~\cite{li2022memorydriven} builds a retrieval image memory banks, then selectively feed image features at each stage of the network. 
Although ~\cite{li2022memorydriven} and our algorithm share the same spirit of utilizing image retrieval, ours doesn't require the multi-stage discriminators~\cite{li2022memorydriven, Xu18} to generate high-resolution images.

\noindent\textbf{Joint vison-language representation.}
Vision and language (VL) methods~\cite{DBLP:conf/iccv/GuWCC17,DBLP:conf/cvpr/KarpathyL15,DBLP:journals/corr/KirosSZ14,DBLP:conf/cvpr/GuCJN018,DBLP:conf/cvpr/WangLL16,DBLP:conf/nips/LuBPL19,DBLP:conf/emnlp/TanB19,DBLP:conf/eccv/ChenLYK0G0020,li2019visualbert,gu2020self} are representatives that embrace multi-modal information for many computer vision tasks, such as image captioning and cross-modal retrieval.
Such methods aim at mapping text and images into a common space, where semantic similarity across different modalities can be learned by ranking-based contrastive losses.
DAMSM~\cite{Xu18} computes the similarity between images and captions, which has been widely used in text-to-image generation methods~\cite{Xu18, hinz2020semantic, zhu2019dm} for additional and fine-grained feedback.  
CLIP~\cite{DBLP:conf/icml/RadfordKHRGASAM21} provides more powerful joint representation, which is pretrained by scaling up the simple language and image proxy task.
It achieves great success in inference-time generation re-ranking~\cite{DBLP:conf/icml/RameshPGGVRCS21} or text-driven image manipulation~\cite{DBLP:conf/iccv/PatashnikWSCL21}.
Our method conducts the offline cross-modal search using image-text representations from~\cite{Xu18}, and has the flexibility to incorporate with other joint vison-language pre-trainng models.

\noindent \textbf{Hypernetwork and its application in image synthesis.}
Hypernetwork~\cite{DBLP:conf/iclr/HaDL17} is proposed to predict the model parameters for a target network and has been used in image synthesis tasks~\cite{inr_gan,anokhin2020image,haydarov2022hypercgan,Chiang_2022_WACV}.
~\cite{inr_gan} presents impressive ability in unconditional image generation by modulating the generators using hypernetworks.
HyperCGAN~\cite{haydarov2022hypercgan} extends these methods to text-to-image generation task by utilizing the text-conditional hypernetwork-based image generator.
~\cite{Chiang_2022_WACV} learns an implicit representation of the 3D scene with the neural radiance fields model (NeRF)~\cite{DBLP:conf/eccv/MildenhallSTBRN20}, in which a hypernetwork is used to transfer the style information.
Our method also uses representative modulation power of hypernetwork but only applies it to the encoding scheme for transferring the text-conditional visual information (e.g. content, layout) into latent domain.

\section{Method}

\begin{figure}[h]
\begin{center}
\centerline{\includegraphics[width=\columnwidth]{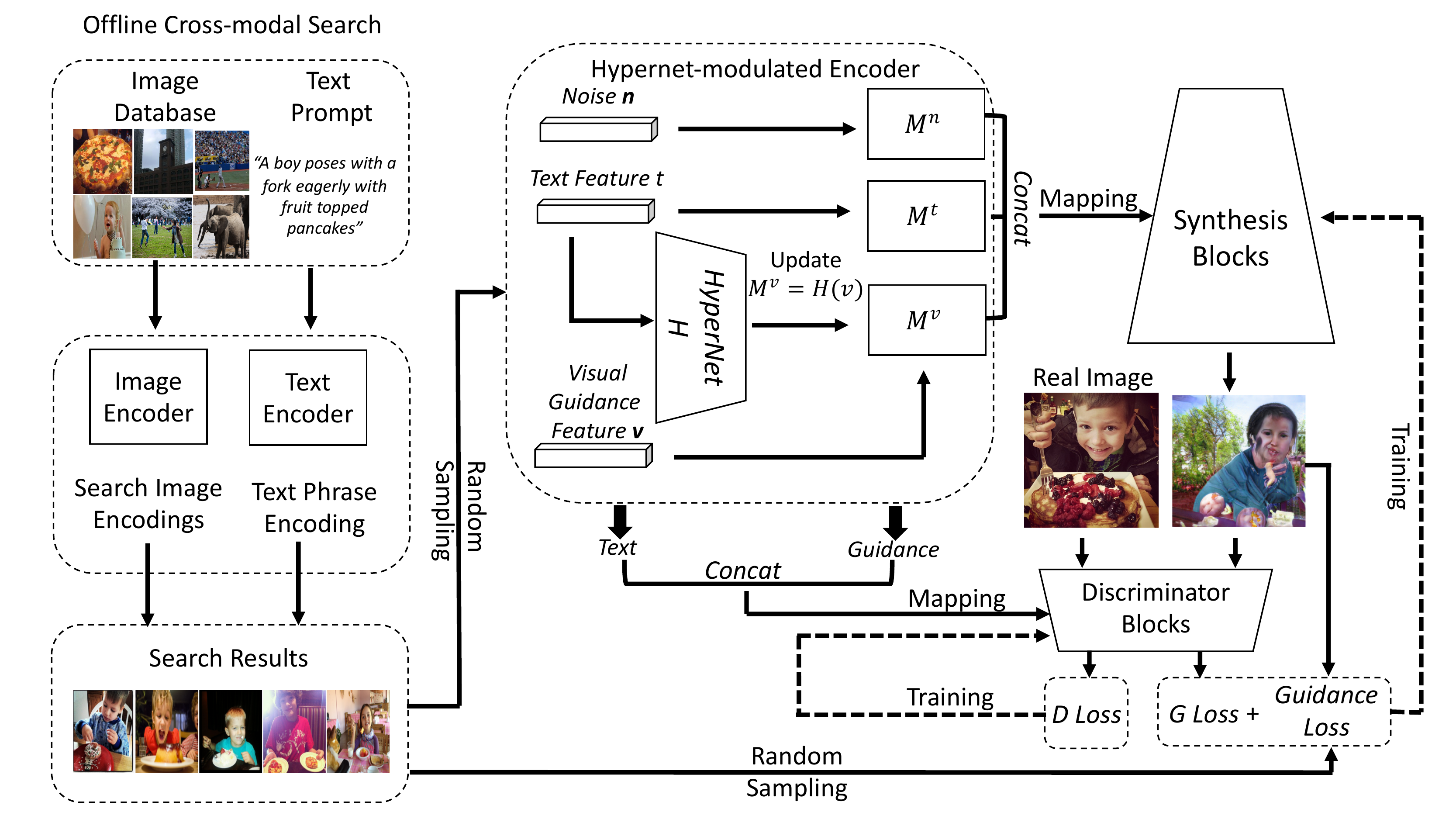}}
\caption{System overview: The proposed framework consists of two components \textit{left}: offline cross-modal search and \textit{right}: unified training accommodating multi-modal data stream.
The cross-modal is designed to build up an offline mapping between captions and image data with high semantic correlation in the embedding space. Dynamic image-text pairs are fed through random sampling to the novel hypernet-modulated visual-text encoder, which are instantiated separately for both generator and discriminator. 
To be specific, hypernetworks take input of text information to predict the weights of visual encoding module $\bM^v$, which transfer the visual information from visual guidance features into the latent representation. All modules are end-to-end trainable, which are optimized by the combination of guidance loss and GAN loss via back propagation.
}
\label{fig:framework}
\end{center}
\end{figure}

Figure~\ref{fig:framework} shows the overall architecture for the proposed training framework. 
The system is composed of two schemes: 
offline cross-modal search and unified training with dynamically created image-text pairs.
The cross-modal search is designed to build up an offline mapping between captions and the image database with high semantic correlations.
The mapping helps to feed random data streaming continuously and efficiently into the training stage.
The training scheme combines the generative power of StyleGAN architectures~\cite{DBLP:conf/cvpr/KarrasLA19, Karras2019stylegan2} with additional data samples provided by text and corresponding images in an end-to-end manner.
The original mapping layer in StyleGAN lacks the ability of encoding semantic information from higher-level concepts into intermediate style representation (e.g. text) during training. 
We address this issue by introducing a new encoding scheme fitting for different types of input. The encoding is guided by both generative objectives and implicit visual guidance measured using simple feature distances.
After completing the unified training, the generator can be directly used to generate images adhere to characteristics described by text and image data.
\subsection{Cross-modal Image Search}
To use rich visual information for better diversity and controllability during training, we build up a candidate image database which are highly correlated with text descriptions via cross-modal image search.
We first embed the representation of images and captions into a common space and then use the cosine similarity as ranking metrics. 
We denote the text set $D_C=\{{C_i}\}_{N_C}$ with $N_C$ captions and image database as $D_I=\{{I_j}\}_{N_I}$ with $N_I$ images. 
We generate the visual feature vectors using image encoder $E_I$ and text features using text encoder $E_C$ in the common feature space:
\begin{eqnarray} \label{feat_ic}
\bv_i = E_I(I_j, \Theta_{E_I}) \\
\bt_j = E_C(C_i, \Theta_{E_C})
\end{eqnarray}
where $\Theta_{E_I}$ and $\Theta_{E_C}$ are the pre-trained weights of the image and text encoder, respectively. 
We build up a similarity score matrix $S = \{s_{ij}\}_{N_C \times N_I}$, where each element reflects the semantic similarity between a query text feature $\bt_i$ and a candidate image feature $\bv_j$.
We thus formulate the image-caption cosine similarity score as:
\begin{eqnarray}
s_{ij} = \frac{\bt_i \cdot \bv_j}{|\bt_i||\bv_j|} 
\end{eqnarray}
Thus we can easily build the mappings $\Gamma=\{\gamma_{ik}\}_{N_C \times K}$ from the $i$th query to search results by recording the corresponding index of visual features with largest $K$ similarity scores.
The offline mappings are beneficial for continuous and efficient data streaming during training. More importantly, comparing with fixed image-text pairs provided by the training dataset itself, it yields much wider variety of pairs for GANs distribution learning.

\subsection{HyperNetwork modulated Visual-Text Encoding}
We start by developing an extension to StyleGAN2~\cite{Karras2019stylegan2} for modeling joint distributions of multimodal data. 
One simple design is adding two separate linear embedding layers $M^t$ and $M^v$ which directly encodes images and text features as follows:
\begin{eqnarray} \label{eq:simple_encoding}
\bv_e = M^v(\bv, \Phi_{M^v}) \label{eq:simple_encoding:visual}\\
\bt_e = M^v(\bt, \Phi_{M^t}) \label{eq:simple_encoding:text}
\end{eqnarray}
where $\bv$, $\bt$ are the pre-trained image and text representation. $\Phi_{M^v}$ and $\Phi_{M^t}$ are trainable parameters.
Then $\bv_e$ and $\bt_e$ concatenated with the noise embedding are encoded to latent domain by additional mapping layers of generator.
The encoding scheme is similarly instantiated for the discriminator except that only text-visual feature are mapped to latent codes as conditions. 
However, the separate embedding layers fail to further encode the image and text correlation, which is a  bottleneck in transferring the text-conditioned visual information into the joint latent domain.
Moreover, joint encoding multimodal data without any constraint may also result in uncontrollable distribution coverage for the generator.
One solution is to directly minimize the distance of $v_e$ and $t_e$. However, the constraint is too strong for such simple linear layers, especially when $v$ and $t$ already present the semantic-correlation in the pretrained cross-modal common space. This may lead to a trivial solution and prevents the latter mapping layers to learn useful joint latent codes.

To address this issue, we use the hypernetwork modulation to transfer the text-conditional visual information (e.g. layout) into latent domain.
Instead of directly optimizing $\Phi_{M^v}$, we first encode $\bt$ with a hypernetwork $\mathcal{H}$ into a feature vector, which is reshaped as a trainable parameter of $M^v$.
We formulates the weights prediction by modifying Eq.~\ref{eq:simple_encoding:visual} as:
\begin{eqnarray} \label{eq:hyper_encoding}
\bv_e = M^v(\bv, \Phi_{M^v}=\mathtt{reshaped}(\mathcal{H}(\bt)))
\end{eqnarray}
where $\mathcal{H}$ takes text information as input and predicts the weights updating of $\bM^v$, i.e. $\Phi_{M^v}$.
In our experiments, we report the performance with hypernetwork modulation and also conduct investigation on the direct optimization.


\subsection{Optimization with Implicit Visual Guidance}
We train the generator $G$ and discriminator $D$ driven by the combination of generative objectives and implicit visual guidance from retrieval images.
We first accommodate the conventional generator loss with multimodal condition (i.e., additional information of caption and corresponding search images):
\begin{eqnarray}
\mathcal{L}_{gen} &=&\mathbb{E}_{z \sim p(z), k_1 \sim \Gamma_i}[\log(1- \\ \nonumber
&&D(G(z, \bt_e^{(i)}, \bv_e^{(k_1)}), \bt_e^{(i)}, \bv_e^{(k_1)})]
\end{eqnarray}
where $z$ is sampled from normal distribution $N(0,I)$. $\Gamma_i$ is the offline mapping for the $i$th query caption. We randomly sample $k_1$ to dynamically formulate an image.
$t_e^{i}$ and $v_e^{k_1}$ are derived using Eq.~\ref{eq:simple_encoding:text} and Eq.~\ref{eq:simple_encoding:visual} and are trained together with the proposed hypernetwork-modulated encoding scheme in an end-to-end manner.

To implicitly exploit the visual information, we use the visual guidance loss measured using the distance between the generation result and the reference image:
\begin{eqnarray}
\mathcal{L}_{guide} = \mathbb{E}_{z \sim p(z), k_1, k_2 \sim \Gamma_i}||E_I(G(z, \bt_e^{(i)}, \bv_e^{(k_1)})) - \bv_e^{(k_2)}||
\end{eqnarray}
where $E_I$ is a pre-trained image encoder used in our cross-modal search system.
Our final loss function for generator is
\begin{eqnarray}
\mathcal{L}_G = \mathcal{L}_{gen} + \lambda \mathcal{L}_{guide}
\end{eqnarray}
where we set $\lambda$ as 1.0 in the implementation.
We similarly adapt the discriminator by incorporating caption and image searching and minimize
\begin{eqnarray}
\mathcal{L}_D = &&-\mathbb{E}_{k_1 \sim \Gamma_i} [\log D(I^{(i)},\bt_e^{(i)}, \bv_e^{(k_1)})] - \\ \nonumber
&&\mathbb{E}_{\hat{x} \sim p(G), k_1 \sim \Gamma_i}[\log(1-D(\hat{I}^{(i)}, \bt_e^{(i)}, \bv_e^{(k_1)})]
\end{eqnarray}
where $\hat{I}^{(i)}$ is the generation result from $G$.
We then alternatively optimize $G$ and $D$ under both conditional and implicit control in the unified training framework.

\section{Experiments}
\begin{table}[tb]
\footnotesize
\begin{center}
\caption{FID Comparisons on CUB dataset.}
\label{tab:cub:result}
\begin{tabular}
{lccccccccc}
\toprule
&\multicolumn{1}{c}{{Methods}}
&\multicolumn{1}{c}{FID $\downarrow$} \\
\midrule
&Attn-GAN~\cite{Xu18} &23.98\\
&ControlGAN~\cite{DBLP:conf/nips/LiQLT19} &13.92  \\
&DM-GAN~\cite{zhu2019dm} &16.09 \\
&DF-GAN~\cite{DBLP:journals/corr/abs-2008-05865} &14.81  \\
&Memory-GAN~\cite{li2022memorydriven} &\underline{10.49} \\
\midrule
&Ours &\textbf{5.65}\\
\bottomrule
\end{tabular}
\end{center}
\end{table}


\begin{table*}[tb]
\footnotesize
\begin{center}
\caption{Comparisons with the state-of-the-art methods. Note that Ours-DAMSM and Ours-CLIP refer text-image representations pre-trained by DAMSM~\cite{Xu18} and CLIP~\cite{DBLP:conf/icml/RadfordKHRGASAM21}, respcetively.}
\label{tab:coco:result}
\begin{tabular}
{lccccccccc}
\toprule
&\multicolumn{1}{c}{{Methods}}
&\multicolumn{1}{c}{Params (M) $\downarrow$}
&\multicolumn{1}{c}{IS$ \uparrow$}
&\multicolumn{1}{c}{FID $\downarrow$} 
&\multicolumn{1}{c}{SOA-C $\uparrow$}
&\multicolumn{1}{c}{SOA-I $\uparrow$} \\
\midrule
&Attn-GAN~\cite{Xu18} &\underline{$13$} &$23.61\pm 0.21$ &$33.10 \pm 0.11$ &$25.88$ &$39.01$\\
&ControlGAN~\cite{DBLP:conf/nips/LiQLT19} &- &$24.06\pm 0.60$ &-  &- &-\\
&Obj-GAN~\cite{objgan19} &$34$ &$24.09\pm 0.28$ &$36.52 \pm 0.13$  &$27.14$ &$41.24$ \\
&DM-GAN~\cite{zhu2019dm} &$21$ &$\textbf{32.32} \pm \textbf{0.23}$ &$27.34 \pm 0.11$ &$33.44$ &$48.03$ \\
&OP-GAN~\cite{hinz2020semantic} &$18$ &$27.88 \pm 0.12$ &$24.70 \pm 0.09$  &$35.85$ &$50.47$\\
&DF-GAN~\cite{DBLP:journals/corr/abs-2008-05865} &\textbf{12} &- &$19.32$  &- &- \\
&HyperCGAN~\cite{haydarov2022hypercgan} &- &$21.05$ &$20.81$ &- &-\\
&Memory-GAN~\cite{li2022memorydriven} &- &- &$19.47$  &- &- \\
&XMC-GAN~\cite{DBLP:conf/cvpr/0010KBLY21} (96$\times ch$) &$90$ &$30.45$ &$9.33$  &\textbf{50.94} &\textbf{71.33}\\
&XMC-GAN~\cite{DBLP:conf/cvpr/0010KBLY21} (64$\times ch$) &$43$ &\underline{$30.66$}  &$11.93$  &$39.85$ &$59.78$\\
\midrule
&Ours-DAMSM &$25$ &$29.33 \pm 0.20$ &$\textbf{9.13} \pm \textbf{0.07}$ &$35.53$ &$50.66$\\
&Ours-CLIP &$25$ &$29.02 \pm 0.23$ &$\underline{9.23 \pm 0.05}$  &\underline{$43.75$} &\underline{$63.23$}\\
\bottomrule
\end{tabular}
\end{center}
\end{table*}
\subsection{Experimental Setup}
We evaluate our model on CUB~\cite{WahCUB_200_2011} and COCO-2014~\cite{DBLP:conf/eccv/LinMBHPRDZ14} datasets. We preprocess CUB following~\cite{Xu18}, train on 8,855 samples and test on 2933 samples, with 10 captions for each. 
For COCO, we use 80k training samples and 40k testing samples, each annotated with 5 captions following~\cite{Xu18}.

\noindent \textbf{Evaluation Metrics.}
Following existing works, we generate 30,000 images with randomly selected unseen text descriptions for validation,
We use Inception Score (IS)~\cite{DBLP:conf/nips/SalimansGZCRCC16} and Fr{\'{e}}chet Inception Distance (FID)~\cite{DBLP:conf/nips/HeuselRUNH17} for generation quality assessment.
In addition to image quality metrics, we also evaluate the semantic alignment between text and generation image.
We adopt Semantic Object Accuracy (SOA)~\cite{hinz2020semantic} to validate whether objects mentioned in the text are recognizable by a YOLO detector~\cite{DBLP:conf/cvpr/RedmonDGF16}.
We report both the SOA-C (measured by the percentage of images per class detects the given object) and SOA-I (the percentage of images a given object is detected). 
We report the number of parameters of the image generator. For hierarchical generation methods which have multiple stages (e.g. shape generator) , we simply report the number of parameters of the last stage generator.

\noindent\textbf{Implementation Details.}
For consistency and fair comparison with most existing text-to-image generation methods, we adopt DAMSM~\cite{Xu18} pretrained image encoder and text encoder. On COCO dataset, we also utilize stronger multimodal representations from CLIP~\cite{DBLP:conf/icml/RadfordKHRGASAM21}. 
During training, we choose top $K=5$ retrieval images for random selection.
During inference, we select the reference image from the training set instead of the validation one for fair comparison.
We use the same set of data augmentation (i.e. random crop and
random horizontal flip) with previous works.
For generation models, we use extensions of StyleGAN2 architectures following~\cite{casanova2021instanceconditioned}. $M^v$ and $M^t$ are both single fully connected layers that encode 256-dim visual or text features into 128-dim for both $G$ and $D$.
The Hypernetwork architecture is implemented as Multilayer perceptrons (MLP)~\cite{DBLP:conf/icml/CollobertB04}, which includes single 64-dim hidden layer with ReLU activation. 
It encodes 256-dim text feature into a large vector, which is reshaped as $256 \times 128$ for $M^v$ weights updating.
We train the models on two 48GB-Nvidia-A40 GPUs for resolutions $256 \times 256$, with learning rate of $3\mathrm{e}{-3}$ for both $G$ and $D$.
We report the results at $300$ epochs when FID doesn't improve during training.

\begin{figure}[tbh]
\begin{center}
\centerline{\includegraphics[width=\columnwidth]{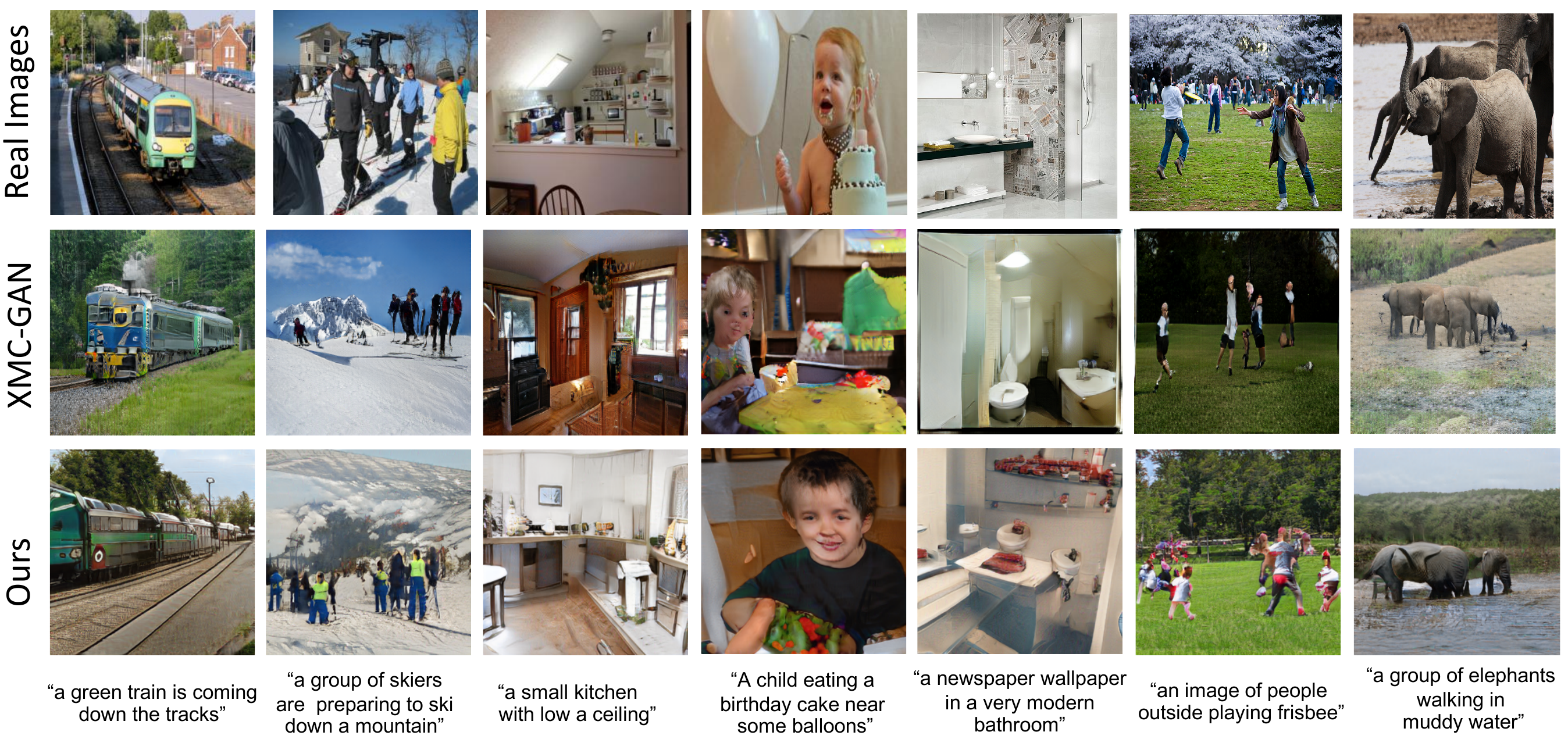}}
\caption{Generation results on COCO. For XMC-GAN, we take samples from original paper~\cite{DBLP:conf/cvpr/0010KBLY21}.
}
\label{fig:visualization}
\end{center}
\vskip -0.4in
\end{figure}
\subsection{CUB Results}
As shown in Table~\ref{tab:cub:result}, our method significantly improves FID to $5.65$, comparing with the leading methods on the CUB dataset.
Our method also outperforms  Memory-GAN method by a large margin, which suggests that ours utilizes the retrieval results in a more effective and efficient way, even without multi-stage discriminators.

\subsection{COCO Results}
\noindent \textbf{Quantitative results.}
We compare our method with representative GAN-based text-to-image generation methods using various metrics. As shown in Table~\ref{tab:coco:result}, we achieve better FID of 9.13 than the state-of-the-art method XMC-GAN with up to $3.5\times$ fewer generator parameters. 
It also validates our main idea: under carefully designed guidance, enlarging diversity of visual-semantic joint distribution is beneficial to the learning of high-quality image generation.
Our method also achieves comparable IS with XMC-GAN. DM-GAN~\cite{zhu2019dm} achieves better IS than both XMC-GAN and ours and can generate more realistic images.
This observation suggests that FID is a more robust metric and IS has limitation in capturing diversity and quality.~\cite{hinz2020semantic}. 

For SOA-C and SOA-I, ours-DAMSM achieves comparable results with OP-GAN. This demonstrates that compared with additional individual boxes, the easily acquired retrieval samples in our unified training framework presents equivalent capability in region-level semantic preservation.
XMC-GAN outperforms all methods by a large margin, showing strong text-alignment ability learnt by text-image contrastive loss, which also suggests the separately pre-trained language model in large-scale (i.e. BERT~\cite{devlin2018bert}) is advantageous over the simple LSTM encoder in DAMSM. 
We also replace the DAMSM with a pre-trained CLIP model to incorporate more powerful joint representations. As shown in Table~\ref{tab:coco:result}, ours-CLIP drastically improves SOA-C and SOA-I and outperforms XMC-GAN($64 \times ch$) even with less generator parameters.

\noindent \textbf{Qualitative Results.}
In Figure~\ref{fig:visualization}, we compare our model with state-of-the-art text-to-image generation method XMC-GAN at resolution $256$ on COCO captions. 
For XMC-GAN, we directly use the generation results from the original paper.
Our model consistently produces more realistic images, especially for human pose and face generation, which are very challenging in complex scenes.

\begin{figure}[tbh]
\begin{center}
\centerline{\includegraphics[width=\columnwidth]{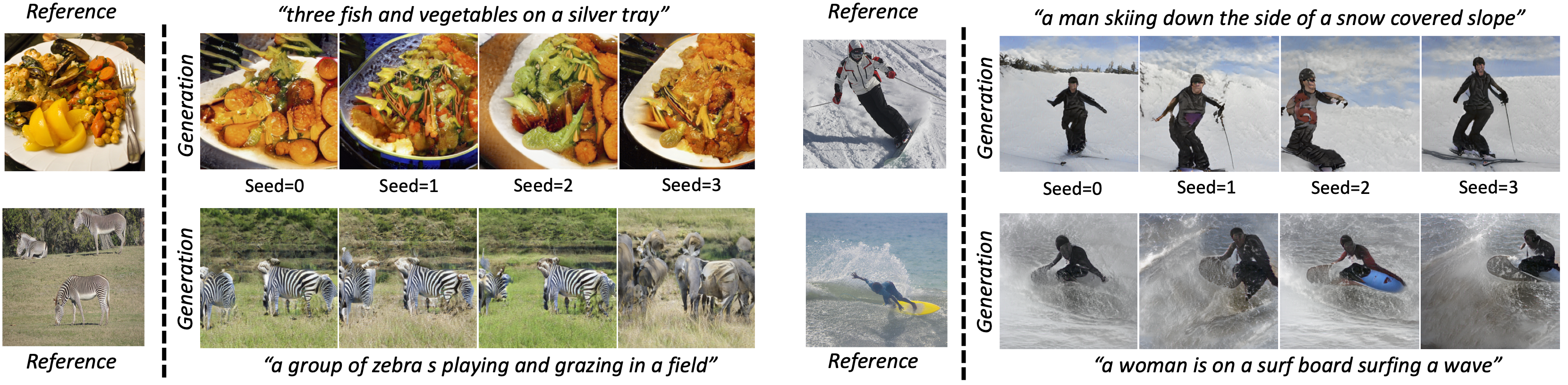}}
\caption{Generation results on COCO with varying noise vectors while fixing retrieval sample.
}
\label{fig:noise_var}
\end{center}
\end{figure}

\begin{figure}[tbh]
\vskip -0.4in
\begin{center}
\centerline{\includegraphics[width=\columnwidth]{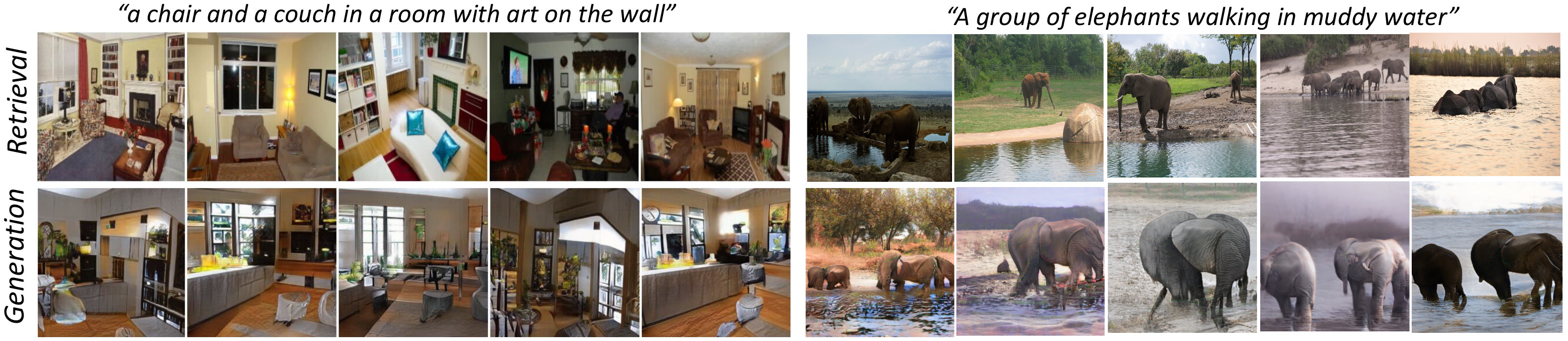}}
\caption{Generation results on COCO with varying retrieval images  while fixing noise. 
}
\label{fig:ref_var}
\end{center}
\end{figure}

\begin{figure}[tbh]
 \begin{center}
   \subfigure[\textit{\textbf{left}:} for the \textit{upper} cases, our generations present close view of traffic lights, which follows the guidance of reference image with sky background; for the \textit{bottom} reference, generations also present a similar layout that traffic lights are in the street view.
\textit{\textbf{right}:} The caption mentions `wildlife' without specifying the species. \textit{upper} reference presents `sheep', which is transferred to generations, showing a group of sheep on grassy area; In contrast, given \textit{bottom} reference of cattle, the generations produce images of cattle.]{
      \includegraphics[width = \columnwidth]{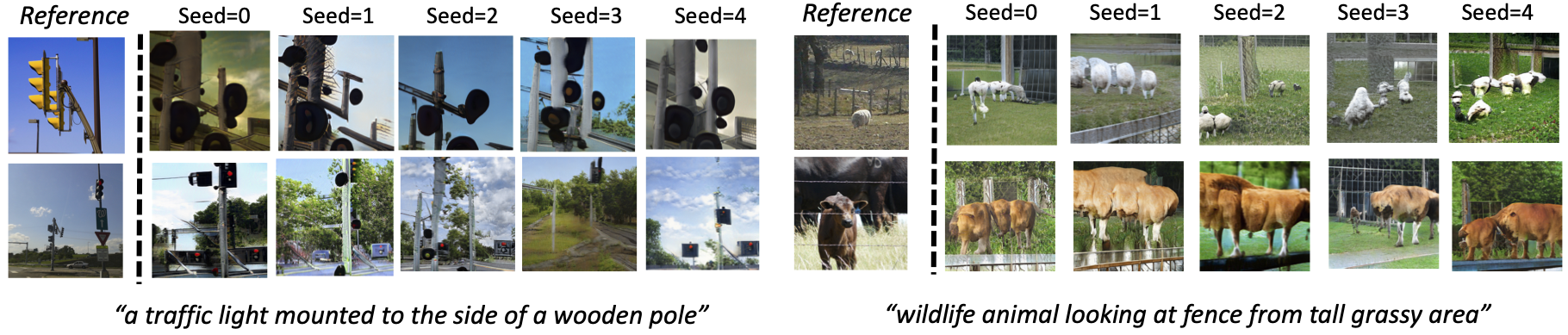}
   }
\end{center}
  \hfill
  \begin{center}
   \subfigure[\textit{\textbf{left}:} both \textit{upper} and  \textit{bottom} cases present a consistent camera view under the guidance of corresponding kitchen images. The layout information in the kitchens is transferred to the generation results.
\textit{\textbf{right}:} \textit{upper} presents fewer passengers in the boat while the \textit{bottom} shows multiple passengers. ] {
      \includegraphics[width = \columnwidth]{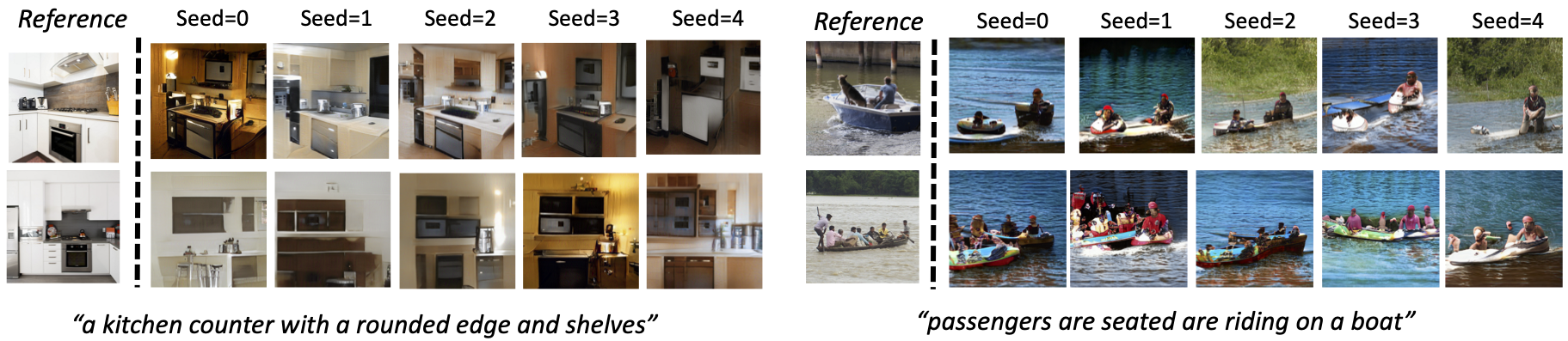}
   }
   \end{center}
  \hfill
  \begin{center}
   \subfigure[\textit{\textbf{left}:} \textit{upper} reference provides the information of two screens, which is also shown in generations results, producing multiple screens on a desk; In contrast, \textit{bottom} only shows single screen.
\textit{\textbf{right}:} \textit{upper} presents three skiers on a mountain while the \textit{bottom} shows two.] {
      \includegraphics[width = \columnwidth]{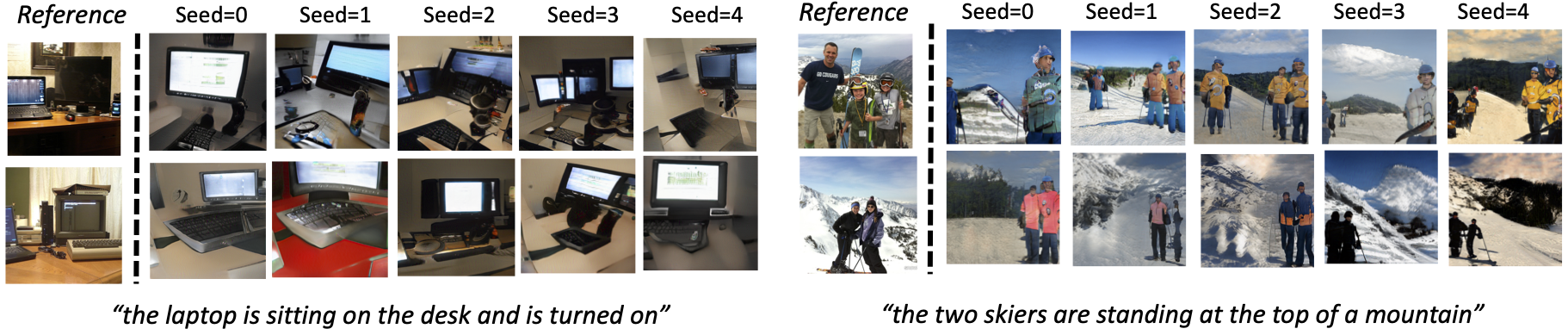}
   }
   \end{center}
   \caption{More qualitative results on COCO varying noise and reference, with clear controllability.
   }
   \label{fig:more_vis}
\end{figure}

\noindent\textbf{Analysis on Diversity and Controllability.}
The dynamic image-text data during training provides not only  additional visual guidance, but also a more diversified joint distribution for the generator to learn a good coverage over.
During testing, the sample diversity can be realized by varying both the noise vector and retrieval images.
We validate this by conducting experiments on COCO under two different settings: 
(1) Generation using the same caption and reference image, but with varying noise vectors. Figure~\ref{fig:noise_var} shows that our method is able to generate diverse images using different noises.
(2) Generation using the same caption and noise vector, but with varying reference image, as shown in Figure~\ref{fig:ref_var}.
We also measure the diversity by calculating the average similarity with $d = \sum_{caps}\sum_{samples}f_{dis}(x_{i}, x_{j})/(C*N)$,
where C and N denotes the number of captions and samples. For $f_{dis}$, we use both $L_2$ and the perceptual distance measured by LPIPS~\cite{DBLP:conf/cvpr/ZhangIESW18}. With both varying noise vectors, we show that further varying reference helps our method increase the distance by $53.2\%$ and $67.8\%$ in terms of $L_2$ and LPIPS, respectively.
Note that both quantitative and qualitative diversity results help us check whether the network simply learns a nearly identity mapping from retrieval samples (i.e., a simple copy-paste operation) or not, which is important but missed by the retrieval-based method MemoryGAN~\cite{li2022memorydriven}. 
Given the diversity of our generations based on either noise or retrieval samples, we conclude that our method learns good distribution instead of a trivial solution. In Figure~\ref{fig:more_vis}, we also observe that the retrieval guidance transfers visual information to generation results in a clear controllable manner, e.g. the backgrounds colors, human poses, room layout, and directions, etc.


\vskip -0.2in
\begin{table*}[tb]
\footnotesize
\begin{center}
\caption{We evaluate the different components on two baselines: Attn-GAN~\cite{Xu18} and Text-conditional StyleGAN2.}
\label{tab:ablation:result}
\begin{tabular}
{lccccccccc}
\toprule
&\multicolumn{1}{c}{{Methods}}
&\multicolumn{1}{c}{Retrieval}
&\multicolumn{1}{c}{HyperNet}
&\multicolumn{1}{c}{Visual Guidance}
&\multicolumn{1}{c}{IS$ \uparrow$}
&\multicolumn{1}{c}{FID $\downarrow$}  \\
\midrule
&Attn-GAN~\cite{Xu18} &\xmark &\xmark &\xmark &23.61 (+0.00) &33.10 (-0.00) \\
&Attn-GAN-Ret &\cmark &\xmark &\xmark &24.40 (+0.79) &34.67 (+1.57) \\
&Attn-GAN~Ret-L1 &\cmark &\xmark &$L_1$ &25.47 (+1.76) &27.30 (-5.80) \\
&Attn-GAN-Ret-Contrast &\cmark  &\xmark &$Contrastive$ &\textbf{27.73 (+4.12)} &\textbf{21.18 (-11.92)} \\
\midrule
&Text-cond StyleGAN2 &\xmark &\xmark &\xmark &20.96 (+0.00) &18.08 (-0.00) \\
&Ours-Ret &\cmark &\xmark &\xmark &21.01 (+0.05) &24.86 (+6.78) \\ 
&Ours-Ret-L1 &\cmark &\xmark &$L_1$ &27.42 (+6.46) &13.87 (-4.21) \\
&Ours-Ret-Contrast &\cmark &\xmark &$Contrastive$ &28.76 (+7.80) &16.35 (-1.73) \\
&Ours-Ret-Hyper-Contrast &\cmark &\cmark &$Contrastive$ &28.43 (+7.47) &10.07 (-8.01) \\
&Ours-Ret-Hyper-L1 (best) &\cmark &\cmark &$L1$ &\textbf{29.33 (+8.37)} &\textbf{9.13 (-8.95)} \\
\bottomrule
\end{tabular}
\end{center}
\end{table*}

\subsection{Ablations}
We evaluate each component by analyzing their effects on two baselines: AttnGAN and text-conditional StyleGAN2. AttnGAN is a representative multi-stage framework adopted by many text-to-image generation methods.
As such, improvements made on it may easily generalize to more.
For StyleGAN2 architectures, we set up the simple baseline by extending it into text-conditional version, based on which we show each component woven together into a unified framework successfully.

\noindent\textbf{Cross-modal search.}
We build offline cross-modal image search for both baselines. As shown in Table~\ref{tab:ablation:result}, both Attn-GAN-Ret and Ours-Ret FID become worse, suggesting that diversify the joint multimodal distribution without control or feedback my confuse the generator learning. Note that the FID degradation in AttnGAN is smaller than StyleGAN2 baseline, suggesting that the additional DAMSM loss can still serves as a guidance between image and text.

\noindent \textbf{Different types of Visual Guidance Loss.}
The implicit visual guidance, which directly applies to output of generator, are varying as L1 or contrastive loss.
As shown in in Table~\ref{tab:ablation:result}, the FID gets consistently improved with both variants, compared with baselines.
We also observe that contrastive loss works better for Attn-GAN while does the opposite for ours. One reason is that the DAMSM loss can be treated as a variant of visual-text contrasive loss, with image-image contrastive samples introduced, the generator can learn both intra- and inter-modal information simultaneously. Another reason is the smaller batch size in our baseline makes it difficult to learn from contrastive samples~\cite{DBLP:journals/corr/abs-2002-05709}

\noindent\textbf{Hypernetwork modulation.}
One issue with Ours-Ret-L1 and Ours-Ret-Contastive is that, the diversity cannot be provided by retrieval samples, i.e., the visual information from retrieval samples cannot be effectively transferred to the latent representation for guidance. We observe that, during the optimization of separate image and text encoding scheme, the weights of the image encoder tend to be suppressed, i.e., instead of learning from diversified joint distribution, it degrades to the simpler one. 
We use the hypernetwork to modulate the image encoding layer to address this issue, enabling effective visual information transfer. Figure~\ref{fig:ref_var}, Figure~\ref{fig:more_vis} and Table~\ref{tab:ablation:result} show diverse generation results and improved FID, demonstrating the effectiveness of hypernetwork modulated encoding scheme.

\subsection{Enhanced Controllability with Run-time Latent Optimization}

Our method is trained in a unified framework and can directly produce generation results with guidance text and image inputs during inference, without any additional optimization (e.g. re-ranking, noise optimization).
Benefit from the diversified joint visual-text distribution provided by dynamic pairs via cross-modal search, the guidance visual information has been successfully learnt and transferred, thus presenting the controllability property \textit{w.r.t.} retrieval guidance images. We further show this controllablity can be enhanced via run-time latent optimization~\cite{styleencoder,DBLP:conf/iccv/AbdalQW19}.

\noindent \textbf{Implementation Details.}
Given a text and reference image pair, we randomly sample 10k images with varying noise vectors, from which we calculate the average and standard deviation of W latent space. We optimize latent code $w_{opt}$ using the perceptual loss~\cite{DBLP:conf/cvpr/ZhangIESW18} between generation result $G_{synth}(w_{opt})$ and reference image $I_{ref}$, i.e. $L_{percept}(G_{synth}(w_{opt}),I_{ref} )$. We use Adam~\cite{KingmaB14}, with $lr=0.02$, $(\beta_1, \beta_2)=(0.9, 0.999)$ for 300 iterations for each image.

\begin{figure}[tbh]
\begin{center}
\centerline{\includegraphics[width=\columnwidth]{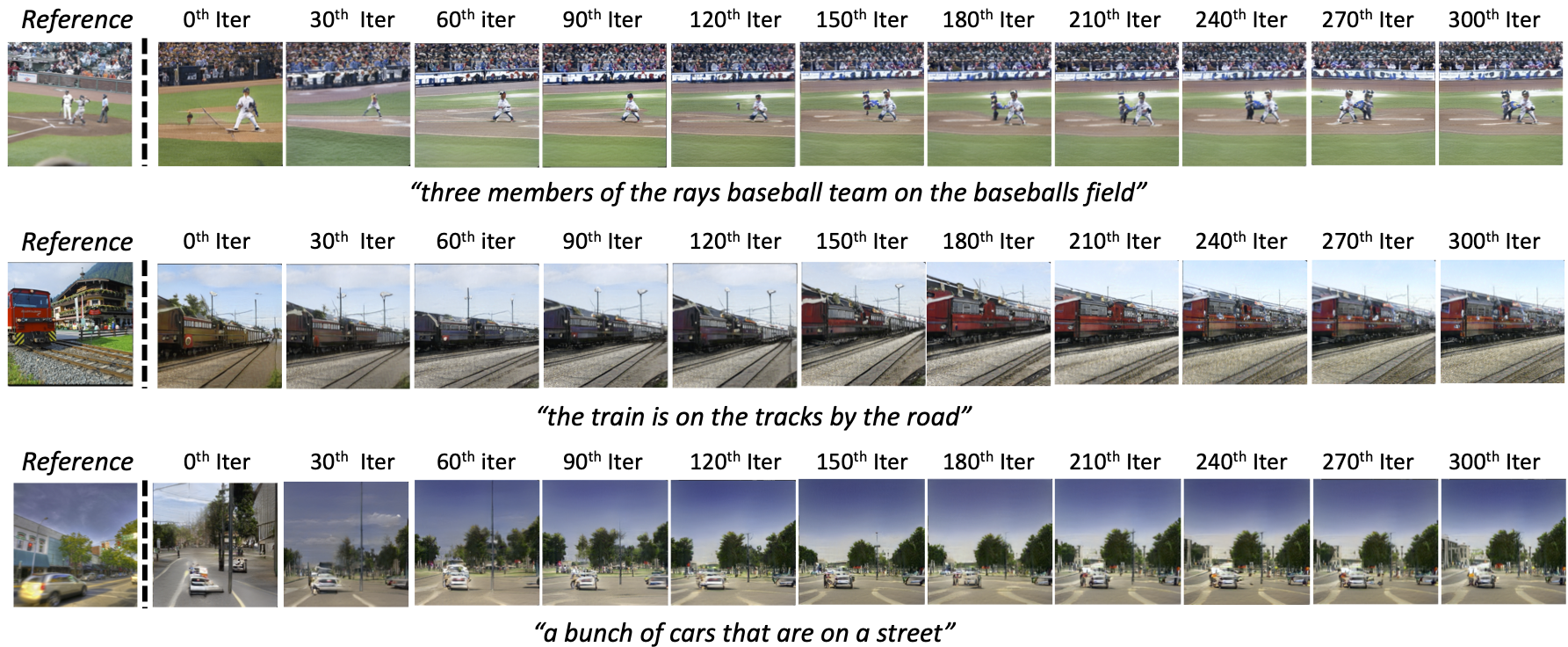}}
\caption{As optimization iterations increases, the network gradually produces similar tones, saturation, layouts, etc.
}
\label{fig:noise_opt_all}
\end{center}
\end{figure}

\noindent\textbf{Results.}
As shown in Figure~\ref{fig:noise_opt_all}, the initial generation (0th iteration) from the original latent space (with averaging 10k samples) and reference images may have large distinctions. After 30 iterations, the style (e.g. tone, color, saturation) starts matching with the guidance. For longer iterations, more detailed information, especially the content and layout can be learnt.

\section{Conclusion}
We propose a new text-to-image generation method that produces high-quaility and diversity images, with controllability provided by easy acquired retrieval images.
Though our method has several components, it is not a collection of orthogonal innovations.
Rather, these components are designed and woven together for a high-level vision: improving generator by training over diversified image-text joint distribution by utilizing retrieval images. 
To facilitate the controllable and effective visual information transfer, we propose the implicit visual guidance and hypernetwork modulated encoding scheme in a unified training framework. Quantitative and qualitative results on CUB and COCO, together with thorough ablations and analysis demonstrate the effectiveness of our method. We also show the enhanced controllability by adopting run-time latent optimization.
For the current system,  retrieval samples are processed as feature vectors for storage and efficiency consideration, trading off with capacity of visual information.
Future work can be done by incorporating with local visual information (e.g. regions), external database for better generations.


\bibliography{aaai23}

\end{document}